# Evolutionary Deep Nets for Non-Intrusive Load Monitoring

Jinsong Wang and Kenneth A. Loparo
Case Western Reserve University, Cleveland, Ohio, USA
jxw741@case.edu; kal4@case.edu

***Abstract*—Compact appliance classifiers can support local electricity-load monitoring where memory and computation are constrained. We evaluate adaptive sparse evolutionary training and warm-start compression using synthetic aggregate windows derived from household appliance recordings. A completed convolutional network study trains 120 models over four targets and three seeds with validation-only model selection. Two analyses examine compression with approximately preserved binary accuracy and joint improvement in positive-class F1 and sparsity. Relative to a newly trained fixed sparse evolutionary baseline, the selected compressed policy changes accuracy from 91.53%, while increasing macro F1 to 92.40%. Stored weights decrease by 17.36% and active first-layer connections by 23.93%. Refinement requires additional training. Three seeds and unverified source-window independence prevent claims of statistical equivalence or unseen-house performance.**

*Index Terms—Demand-side management, load monitoring, neural networks, sparse matrices.*

## I. Introduction

Appliance-level information can help characterize household electricity demand and support demand-side analysis. Non-intrusive load monitoring (NILM) infers appliance behavior from aggregate measurements; neural methods can learn relevant signal features [1]. For local monitoring, a useful model must balance predictive quality with the size of the deployed network. Here, the task is binary appliance-presence classification in one-hour windows, rather than reconstruction of appliance power at every instant. The experiments use synthetic mixtures derived from UK-DALE [2].

Sparse evolutionary training (SET) changes connections during optimization [3]. LASSO shrinks individual weights [4], while group LASSO promotes group selection [5]. Structured regularization extends this idea to neural networks [6]. Gradient-guided connectivity and structured sparse training provide additional mechanisms [7], [8]. Adaptive sparsity is also established in time-series modeling [9]. Our contribution is a measured, reproducible comparison of established components for appliance monitoring, with explicit separation of nonzero connections, stored weight arrays, and dense-kernel computation. We do not claim a new regularizer or demonstrated grid-level operational benefit.

The previous works, including LASSO, group LASSO and LASSO+SET; selected results appear in Table I. Its 288 runs span a feedforward deep neural network (DNN), a convolutional neural network (CNN), four targets and three seeds. The completed accuracy study uses longer training and a different checkpoint rule. It therefore supplies a separate comparison, not a rerun or equal-budget replacement of the pilot.

TABLE I. SELECTED HISTORICAL PILOT RESULTS AT TEN EPOCHS

| Method | DNN macro F1 (%) | CNN macro F1 (%) | CNN weights (KiB) |
|---|---|---|---|
| Dense | 80.08 ± 0.28 | 89.06 ± 0.28 | 474.55 |
| LASSO | 79.89 ± 0.35 | 88.96 ± 0.44 | 474.55 |
| Group LASSO | 80.13 ± 0.56 | 88.87 ± 0.61 | 474.55 |
| Fixed SET | 79.71 ± 0.59 | 89.49 ± 0.50 | 474.55 |
| LASSO+SET | 79.36 ± 0.52 | 89.32 ± 0.25 | 474.55 |
| Adaptive group SET | 79.34 ± 0.54 | 89.17 ± 0.31 | 372.96 |

[a] F1 entries are mean ± sample seed standard deviation (SD). Pilot results were not recomputed after implementation corrections. Adaptive group SET reduces CNN stored weights by 21.4% relative to fixed SET, with a 0.32-percentage-point mean F1 loss. This establishes a historical size–quality trade-off, not identical accuracy.

## II. Data and methods

### *A. Data and model*

The supplied data contain 10,000 aggregate windows, each with 600 six-second samples, and four binary label columns. Before deduplication, rows 0–1999 form the test split, rows 2000–3599 the validation split, and the remainder the training split. Exact-input deduplication, retaining test rows before validation and training rows, leaves 5,770 training, 1,474 validation and 1,863 test windows. No conflicting labels occur among identical inputs. Training-only global mean and standard deviation scale every partition. Different mixtures may reuse overlapping source recordings; household and time independence are unverified. The intended appliance order is washing machine, dishwasher, refrigerator and microwave, but generator-backed label verification is absent. We therefore use identifiers T0–T3.

Each target has a separate classifier. The convolutional neural network (CNN) has two 16-filter convolutions with kernel sizes 5 and 3 and stride 2, batch normalization, max pooling, dropout 0.2, hidden widths 100 and 20, and a sigmoid output. The flattened representation has 1,184 features. Sparsity affects the first hidden layer. Dense width controls replace its 100 neurons with 80 or 64; other layers remain unchanged.

*B. Adaptive sparsity and classical controls*

Fixed SET uses the initialization probability min(1, 20(d+n)/(dn)) for d inputs and n neurons, then prunes and randomly regrows 30% of active connections, treating positive and negative weights separately. Adaptive variants adjust the connection budget from validation binary cross-entropy (BCE) and can rank new edges by training gradients. The historical controller requests 10% contraction after two accepted checks, or 10% expansion after two rejected checks, within 2–60% of possible connections. A check is accepted when BCE is at most 1.01 times the preceding best BCE. Structured variants rank neurons by incoming weight-plus-bias norm multiplied by outgoing weight norm.

For the LASSO and group-LASSO controls, w denotes a first-layer weight, g its incoming-neuron weight-and-bias group, η the optimizer learning rate, and $\lambda_1$ and $\lambda_2$ the shrinkage strengths. Their post-update shrinkage rules are

$$w \leftarrow sign(w)\, max(|w| - \eta\lambda_1, 0). \quad (1)$$

$$g \leftarrow max(1 - \eta\lambda_2 \,/\, max(\|g\|_2, 10^{-12}), 0)\, g. \quad (2)$$

Masks are enforced after optimization. The shrinkage steps follow Adam [10] and are not exact proximal steps under its adaptive metric. The historical settings are $\eta = 0.001$, $\lambda_1 = 0.0001$ and $\lambda_2 = 0.001$. LASSO variants remain in the adaptive workflow but were not rerun in the new CNN study.

*C. Accuracy refinement and deployment*

The new study includes dense and fixed-SET sources; reduced-width dense models; gradually pruned students; distilled students; and warm-start refinements. We adapt gradual magnitude pruning [11] to joint neuron and connection budgets. Training uses 15 dense epochs, cubic monotone pruning during epochs 16–35, and 15 epochs with fixed topology. Final widths are 80 or 64 with 25% edge density within retained neurons. For normalized pruning progress r from 0 to 1, a width or density budget follows $b(r) = b_{\text{final}} + (b_{\text{start}} - b_{\text{final}})(1 - r)^3$; widths are rounded to integers. Following temperature-scaled knowledge distillation [12], our binary adaptation mixes supervised BCE (weight 0.5) with the temperature-squared Bernoulli Kullback–Leibler divergence (weight 0.5). Teacher probabilities use training inputs only; temperature is 2.

Warm-start refinement selects the stronger source recipe per target by mean validation F1 across seeds; fixed SET wins for every target. It restores each source checkpoint and zero mask, then fine-tunes for 25 epochs. Cubic pruning during the first 15 epochs reaches width 80, 90 or 96 with at most 20% retained-layer edge density; the last ten epochs freeze topology. Frozen-source distillation uses coefficient 0.25 and temperature 2. Native export removes dead neurons and corresponding next-layer weights. Remaining unstructured zeros still occupy dense arrays; fewer nonzeros alone do not imply faster dense inference.

## III. Experimental protocol and results

The new study comprises 84 initial runs, 24 width-80/90 refinements and 12 width-96 refinements: four targets and seeds 11, 23 and 37. Initial training uses 50 epochs, batch size 128 and Adam [10] with cosine decay [13] from 0.001 to 0.00002, without warm restarts. Refinement adds 25 epochs at 0.0003 to 0.000006. Distilled students add 50 teacher epochs. Table II reports total source/teacher plus student epochs, not equal-compute comparisons.

Checkpoints maximize validation positive-class F1; ties prefer lower validation BCE, then earlier epoch. Thresholds maximize validation F1 over 0.05–0.95 in steps of 0.01; ties within $10^{-12}$ prefer the value closest to 0.5, then the lower value. Pruned checkpoints are eligible only at the final budget. Per-target recipes maximize mean validation F1 across seeds; ties prefer smaller stored weights, then method name. Refinement was a validation-motivated extension, and width 96 was added after validation sensitivity for T2. Both policies were frozen before test evaluation. They reuse trained models and are not additional runs.

TABLE II. COMPLETE NEW CNN COMPARISON ON THE HELD-OUT TEST SPLIT

| **Method** | **Macro F1 (%) mean ± SD** | **Accuracy (%)** | **Weights (KiB)** | **First-layer edges** | **MACs ($10^3$)** | **Train epochs** |
|---|---|---|---|---|---|---|
| Dense teacher 100 | 91.35 ± 0.48 | 89.8059 | 474.55 | 118,400 | 257.92 | 50 |
| Fixed SET 100 | 92.41 ± 0.38 | 91.5325 | 474.55 | 25,680 | 257.92 | 50 |
| Dense 80 | 91.33 ± 0.21 | 89.7522 | 380.41 | 94,720 | 233.84 | 50 |
| Dense 64 | 91.37 ± 0.36 | 89.7835 | 305.10 | 75,776 | 214.58 | 50 |
| Gradual pruning 80 | 91.58 ± 0.26 | 90.2532 | 380.41 | 23,680 | 233.84 | 50 |
| Pruning + KD 80 | 91.65 ± 0.24 | 90.4679 | 380.41 | 23,680 | 233.84 | 100 |
| Pruning + KD 64 | 91.53 ± 0.17 | 90.1056 | 305.10 | 18,944 | 214.58 | 100 |
| Refinement 90 | 92.53 ± 0.39 | 91.4654 | 427.48 | 21,312 | 245.88 | 75 |
| Refinement 80 | 92.56 ± 0.34 | 91.5012 | 380.41 | 18,944 | 233.84 | 75 |
| Refinement 96 | 92.56 ± 0.41 | 91.4609 | 455.72 | 22,733 | 253.11 | 75 |
| Initial policy | 91.54 ± 0.20 | 90.1056 | 323.93 | 20,128 | 219.40 | 87.5 |
| Refined policy | 92.59 ± 0.36 | 91.5280 | 392.18 | 19,536 | 236.85 | 75 |

b. KD: knowledge distillation. Macro F1 averages target positive-class F1 within seed, then reports mean ± sample SD across seeds. Target F1 = 2TP/(2TP + FP + FN), where TP, FP and FN count true positives, false positives and false negatives; a zero denominator returns zero. Binary accuracy averages correct target decisions. Policy training costs average over target pipelines. Weight arrays include kernels, biases and batch-normalization statistics; optimizer state, activations and archive overhead are excluded. KiB = 1,024 bytes. MACs count one kernel multiplication and accumulation as one operation, excluding normalization, pooling, activations and bias additions.

### A. *Compression with approximately preserved binary accuracy*

Against new fixed SET, the refined policy reduces compact stored weights from 474.55 to 392.18 KiB (17.36%). Binary accuracy changes from 91.5325% to 91.5280%, a decrease of 0.0045 percentage points; both round to 91.53%. This is an approximately accuracy-preserving observation at the stated precision. No prespecified tolerance or statistical equivalence is established. More aggressively pruned students lose appreciable prediction quality.

### B. *Joint improvement in F1 and sparsity*

Using the same cohort and comparator, macro F1 rises from 92.4065% to 92.5896% (+0.1831 percentage points), active first-layer edges fall from 25,680 to 19,536 (23.93%), and compact dense MACs fall from 257,924 to 236,854 (8.17%). Original-matrix first-layer sparsity rises from 78.31% to 83.50%. The refined policy selects width 80 for T0–T2 and 90 for T3. F1 improves for T0, T1 and T3, but falls from 78.2414% to 78.1775% for T2. All three paired seed macro differences are positive; the exact two-sided sign-flip p value is 0.25 (unadjusted and conditional on these three seed differences). These two analyses are not independent replications, and binary accuracy does not improve.

## IV. TRAINING BEHAVIOR

Fig. 1 shows the measured 50-epoch trajectories. Curves first average targets within each seed, then average the three seeds; shaded regions show sample seed SD. Training BCE includes dropout, while validation BCE is evaluated without dropout. Per-epoch validation F1 uses a validation-selected threshold. Reported test scores use the saved checkpoint rather than necessarily the last epoch.

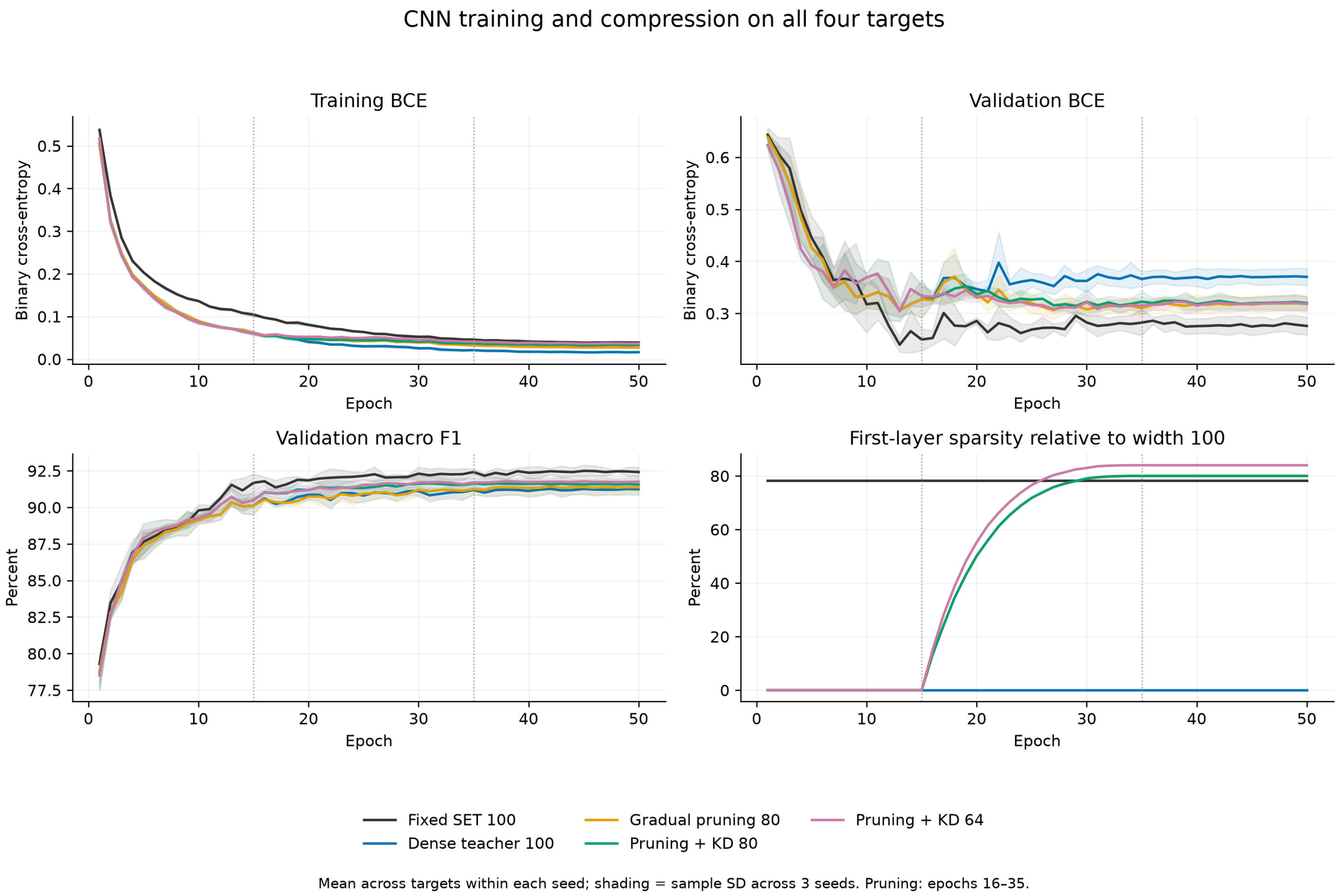

Fig. 1. Initial training and validation trajectories with first-layer sparsity measured relative to width 100. Dotted boundaries identify the gradual-pruning interval. These are validation curves, not test learning curves.

Fig. 2 starts after each selected 50-epoch source checkpoint. The fine-tuning axis therefore excludes source-training cost. Width reduction occurs during the first 15 refinement epochs; topology then remains fixed. Warm starts retain stronger predictive quality than the compressed students trained from scratch in Table II.

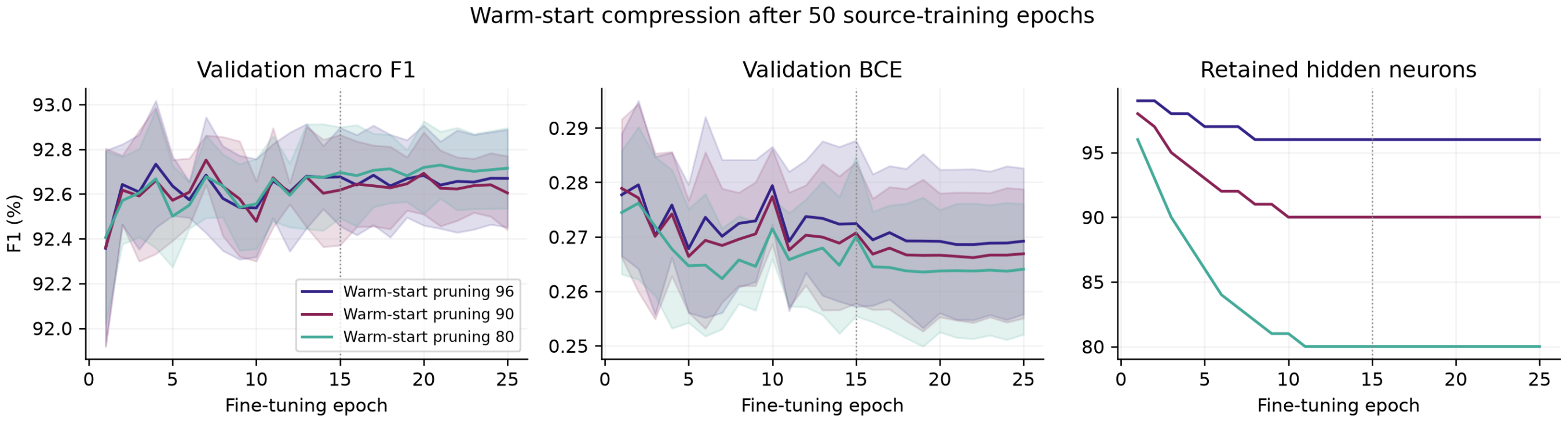

Fig. 2. Twenty-five measured refinement epochs after source training. The three panels show validation macro F1, validation BCE and retained hidden neurons; shading shows sample seed SD.

## V. Discussion and limitations

The results concern synthetic aggregate windows. Exact deduplication cannot exclude overlapping or reused appliance recordings across partitions; household independence and appliance-name provenance remain unverified. Seed SD quantifies training randomness on this dataset, not variation across households. Repeated validation adaptation, including refinement, can overfit that partition.

Extra training and distillation confound the compression effect. A 75-epoch unpruned SET control with matched learning-rate continuation and validation exposure is needed. Further evaluation should match deployed sizes, separate source recordings, use more seeds, and prespecify an accuracy-equivalence or noninferiority criterion. MAC reductions do not establish latency or energy gains.

## VI. Conclusion

The selected warm-start policy reduces stored weights by 17.36% and active first-layer connections by 23.93%. Binary accuracy rounds to 91.53% for both models, while macro F1 rises by 0.1831 percentage points. Additional training, three seeds and unverified source independence limit these findings.

## VII. Code and data availability

Code and summary results are available at https://github.com/jinsongwang/EvoNets.

## VIII. References

[1] J. Kelly and W. Knottenbelt, "Neural NILM: Deep Neural Networks Applied to Energy Disaggregation," in Proc. ACM BuildSys, 2015, pp. 55–64, doi: 10.1145/2821650.2821672.

[2] J. Kelly and W. Knottenbelt, "The UK-DALE dataset, domestic appliance-level electricity demand and whole-house demand from five UK homes," Scientific Data, vol. 2, art. 150007, 2015, doi: 10.1038/sdata.2015.7.

[3] D. C. Mocanu, E. Mocanu, P. Stone, P. H. Nguyen, M. Gibescu, and A. Liotta, "Scalable training of artificial neural networks with adaptive sparse connectivity inspired by network science," Nature Communications, vol. 9, art. 2383, 2018. doi: 10.1038/s41467-018-04316-3.

[4] R. Tibshirani, "Regression Shrinkage and Selection via the Lasso," J. Roy. Statist. Soc. B, vol. 58, no. 1, pp. 267–288, 1996, doi: 10.1111/j.2517-6161.1996.tb02080.x.

[5] M. Yuan and Y. Lin, "Model selection and estimation in regression with grouped variables," J. Roy. Statist. Soc. B, vol. 68, no. 1, pp. 49–67, 2006, doi: 10.1111/j.1467-9868.2005.00532.x.

[6] W. Wen, C. Wu, Y. Wang, Y. Chen, and H. Li, "Learning Structured Sparsity in Deep Neural Networks," in Advances in Neural Information Processing Systems, vol. 29, 2016. [Online]. Available: https://arxiv.org/abs/1608.03665

[7] U. Evci, T. Gale, J. Menick, P. S. Castro, and E. Elsen, "Rigging the Lottery: Making All Tickets Winners," in Proc. ICML, PMLR, vol. 119, pp. 2943–2952, 2020.

[8] M. Lasby, A. Golubeva, U. Evci, M. Nica, and Y. Ioannou, "Dynamic Sparse Training with Structured Sparsity," in Proc. ICLR, 2024.

[9] Z. Atashgahi, M. Pechenizkiy, R. Veldhuis, and D. C. Mocanu, "Adaptive sparsity level during training for efficient time series forecasting with transformers," arXiv:2305.18382, 2023, revised 2024.

[10] D. P. Kingma and J. Ba, "Adam: A Method for Stochastic Optimization," in Proc. ICLR, 2015. [Online]. Available: https://arxiv.org/abs/1412.6980

[11] M. Zhu and S. Gupta, "To prune, or not to prune: exploring the efficacy of pruning for model compression," arXiv:1710.01878, 2017.

[12] G. Hinton, O. Vinyals, and J. Dean, "Distilling the Knowledge in a Neural Network," arXiv:1503.02531, 2015.

[13] I. Loshchilov and F. Hutter, "SGDR: Stochastic Gradient Descent with Warm Restarts," in Proc. ICLR, 2017. [Online]. Available: https://arxiv.org/abs/1608.03983